\documentclass[10pt,twocolumn,letterpaper]{article}

\usepackage{cvpr}              
\definecolor{cvprblue}{rgb}{0.21,0.49,0.74}
\usepackage[pagebackref,breaklinks,colorlinks,allcolors=cvprblue]{hyperref}

\usepackage{url}
\usepackage{booktabs}       
\usepackage{amsfonts}       
\usepackage{nicefrac}       
\usepackage{microtype}      
\usepackage[table]{xcolor}         

\usepackage{amsmath}
\usepackage{graphicx}
\usepackage{enumitem}
\usepackage{makecell}
\usepackage{multirow}
\usepackage{arydshln}
\usepackage{subcaption}
\usepackage{amssymb}
\usepackage{pifont}
\usepackage[accsupp]{axessibility}  
\usepackage{accents}
\usepackage{algorithm}
\usepackage{algpseudocode}
\usepackage{mathtools}

\usepackage{color,soul}

\definecolor{first}{RGB}{255, 172, 172}   
\definecolor{second}{RGB}{255, 208, 142}  
\definecolor{third}{RGB}{255, 255, 173}   

\usepackage{array}
\newcolumntype{$}{>{\global\let\currentrowstyle\relax}}
\newcolumntype{^}{>{\currentrowstyle}}
\newcommand{\rowstyle}[1]{\gdef\currentrowstyle{#1}%
  #1\ignorespaces
}
\newcommand{\flip}{\reflectbox{F}LIP}

\newcommand{\methodname}{Softmax-GS}
\newcommand{\projectpage}{\url{http://arthurhero.github.io/projects/smgs/}}

\def\paperID{****} 
\def\confName{CVPR}
\def\confYear{2026}

\title{Softmax-GS: Generalized Gaussians Learning When to Blend or Bound}

\author{
Chen Ziwen$^1$~~~
Peng Wang$^2$~~~
Hao Tan$^1$~~~ 
Zexiang Xu$^3$~~~
Li Fuxin$^4$~~~\\
\vspace{-0.1in}
\normalsize
$^1$Adobe Research~~~
$^2$Tripo AI~~~
$^3$Hillbot~~~
$^4$Oregon State University\\
}

\begin{document}
\maketitle

\begin{abstract}
3D Gaussian Splatting (3D GS) is widely adopted for novel view synthesis due to its high training and rendering efficiency. However, its efficiency relies on the key assumption that Gaussians do not overlap in the 3D space, which leads to noticeable artifacts and view inconsistencies. 
In addition, the inherently diffuse boundaries of Gaussians hinder accurate reconstruction of sharp object edges.
We propose \methodname{}, a unified solution that addresses both the view-inconsistency and the diffuse-boundary problem by enforcing a softmax-based competition in overlapping regions between two Gaussians. With learnable parameters controlling the strength of the competition, it enables a continuous spectrum from smooth color blending to crisp, well-defined boundaries. Our formulation explicitly preserves order invariance for any two overlapping Gaussians and ensures that the output transmittance remains unchanged irrespective of the extent of overlapping, preventing undesirable discontinuities in the rendered output.
Ablation experiments on simple geometries demonstrate the effectiveness of each component of \methodname{}, and evaluations on real-world benchmarks show that it achieves state-of-the-art performance, improving both reconstruction quality and parameter efficiency.
Project page: \projectpage
\end{abstract}    
\vspace{-0.2in}

\section{Introduction}
\label{sec:intro}

\begin{figure}[h]
    \centering
    \includegraphics[width=0.8\linewidth]{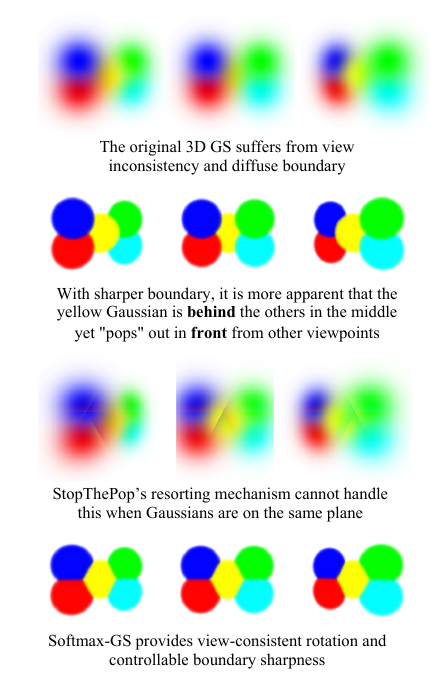}
    \vskip -0.2in
    \caption{\small Comparison between different versions of 3D GS and \methodname{} under slight left and right rotation. Vanilla 3D GS suffers from diffuse boundary and view inconsistency (``popping effect") due to the no-overlap assumption. Only resorting\cite{stopthepop} cannot fix view inconsistency for Gaussians on the same surface. \methodname{} provides both boundary sharpness control and viewpoint consistency between overlapping Gaussians. 360° rotation videos are available on the project page.} 
    \label{fig:smgs_rot}
    \vskip -0.2in
\end{figure}

\begin{figure*}[h]
    \centering
    \includegraphics[width=0.95\linewidth]{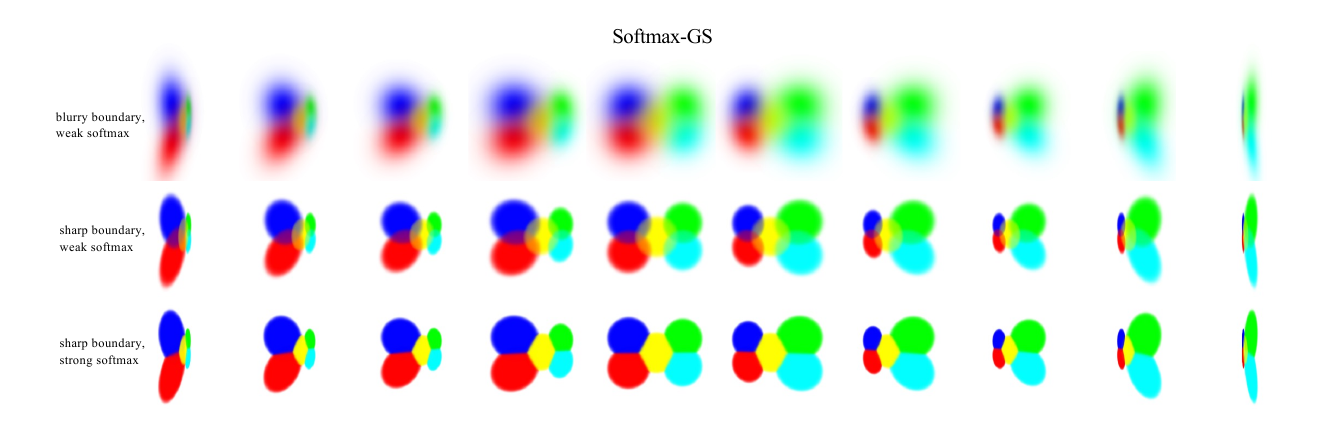}
    \vskip -0.2in
    \caption{\small \methodname{} provides flexible boundary sharpness control and introduces a viewpoint-consistent, softmax-based color-merging mechanism for overlapping Gaussians, enabling both smooth color blending and a winner-take-all behavior.} 
    \label{fig:smgs_comp_main}
    \vskip -0.2in
\end{figure*}

3D Gaussian Splatting (3D GS)~\citep{3dgs} has rapidly gained popularity over Neural Radiance Fields (NeRF)~\citep{nerf} due to its substantially higher efficiency in both training and rendering. 
This breakthrough has sparked many follow-up work  aimed at further improving its capabilities, including advancements in the optimization pipeline~\citep{gsmcmc,sss}, densification strategies~\citep{absgs,revisingdense}, pruning techniques~\citep{pup3dgs,speedysplat}, incorporation of geometric priors and neural networks~\citep{scaffoldgs,sags}, compression for on-device deployment~\citep{compactgs,lightgs,lightweight}, and extensions to large-scale city-level reconstruction~\citep{citygaussian,octreegs,vastgaussian}, among others.

In this paper, we focus on two primary limitations of 3D GS: blurry boundaries and view inconsistency. Due to the long tails of the Gaussian function, 3D Gaussians inherently exhibit blurry boundaries that cannot be directly controlled. As a result, a large number of Gaussians are required to capture sharp color transitions in the input views. Several works~\citep{ges,convexgs,3dhgs,drks,discgs,sss} have addressed this issue by making boundary sharpness adjustable, either through modifications to the Gaussian function or by replacing Gaussians with alternative geometric primitives.

However, simply adjusting the boundary sharpness of individual Gaussians cannot create sharp edges ``between" two overlapping Gaussians in 3D space. 
In fact, the standard 3D GS algorithm assumes that the Gaussians \textbf{never} overlap with each other. This assumption is crucial to its rendering efficiency but also leads to view inconsistency—commonly known as the “popping effect”—when the camera undergoes even slight rotations. To mitigate this, several works~\citep{stopthepop,sortfree,stochasticsplat,hybridtransparency} propose to re-sort the Gaussians based on pixel depth after the standard global sorting step. However, because these methods do not directly challenge the no-overlap assumption—the root cause of the inconsistency—they fail in cases where two flat Gaussians lie on the same surface, i.e., when their pixel-wise depth values coincide (see Fig.~\ref{fig:smgs_rot}). \citet{ever} explicitly recognize this limitation and propose replacing Gaussians with constant-density ellipsoids, deriving the exact density integration along camera rays. While effective, this approach reduces rendering speed by more than an order of magnitude and does not provide controllable boundary sharpness.


We propose \methodname{}, an algorithm that enforces softmax competition among overlapping Gaussians with a tunable sharpness parameter. With some mathematical derivations, we are able to approximate the rendering equation with overlapping Gaussians using the original sorting and splatting approach. To enable individual-Gaussian edge sharpness control, we additionally replace the standard Gaussian kernel with the Generalized Exponential Function (GEF) introduced by GES \citep{ges}. 
By directly relaxing the no-overlap assumption, \methodname{} addresses both the view-inconsistency and blurry-boundary issues in a unified manner, supporting a continuous range of visual effects, from smooth color blending to crisp, well-defined boundaries.

Through optimization experiments on simple geometric color patterns, we demonstrate that neither individual boundary sharpness control nor softmax competition alone is sufficient for optimal performance; both components are essential for fully flexible optimization. We further evaluate \methodname{} on established real-world benchmarks~\citep{mip360,tanks,deepblending}, where it can either \textbf{halve the number} of Gaussians without degrading rendering quality, or alternatively improve reconstruction fidelity while retaining a rendering speed close to that of the original 3D GS. By jointly addressing both the blurry-boundary and the view-inconsistency problem—two fundamental limitations of Gaussian Splatting—\methodname{} provides a principled framework that advances the visual quality of 3D scene reconstruction.

\section{Related Work}
\label{sec:related}

In this section, we summarize prior approaches that address the two major challenges in 3D Gaussian Splatting: the blurry boundary issue and the view-inconsistency issue.

\textbf{Controllable boundary sharpness.}
To address the challenge of controllable boundary sharpness, several approaches have been proposed. GES~\citep{ges} replaces the Gaussian function with the Generalized Exponential Function (GEF), in which the squared distance term $\|\cdot\|^2$ is generalized to $\|\cdot\|^{2\alpha}$, where $\alpha$ is a learnable parameter. However, instead of directly implementing GEF, GES employs an approximate rasterization. In contrast, our method incorporates the exact GEF within our CUDA kernel. SSS~\citep{sss} adopts the Student’s t-distribution as an alternative to the Gaussian. DisC-GS~\citep{discgs} models sharpness by optimizing Bézier curve control points, while 3DCS~\citep{convexgs} replaces Gaussians with convex polygons, incorporating learnable corner smoothness and edge blurriness. DRK~\citep{drks} constructs geometry primitives by connecting offset dots from the centers of the primitives. Finally, 3D-HGS~\citep{3dhgs} enforces sharpness by cutting a Gaussian in half, yielding a sharper representation at the cross-section.
We choose to adopt GEF in this work but most of the above methods can be similarly integrated with the proposed \methodname{} to control the sharpness of Gaussians.

\textbf{View inconsistency.} StopThePop~\citep{stopthepop} highlights the intrinsic “popping” effect in 3D Gaussian rendering, which arises because Gaussians are globally sorted by their center depths. When two Gaussians intersect, even a slight camera movement can abruptly change their relative order, causing sudden changes in the rendered images. StopThePop addresses this issue by recomputing per-pixel depths for the Gaussian splats and re-sorting them for each pixel according to these depths. However, since it retains the non-overlapping assumption of the vanilla 3D GS, it cannot resolve the artifacts when two flat Gaussians lie on the same surface, where their per-pixel depths coincide (see Fig.~\ref{fig:smgs_rot}).
LC-WSR~\citep{sortfree} attempts to eliminate the computationally expensive re-sorting step of StopThePop by learning to predict mixing weights for each Gaussian from the depth values, but the learned weights do not necessarily respect the rendering equation outside the training views. Similarly, StochasticSplats~\citep{stochasticsplat} uses Monte Carlo sampling to mix pixel colors according to the transparency of overlapping Gaussians, demonstrating the ability to blend the colors of intersecting Gaussians. In comparison, our method provides additional control by enabling a smooth transition from normal color blending to sharp boundaries between overlapping Gaussians. 
Finally, EVER~\citep{ever} explicitly addresses the non-overlapping assumption by replacing 3D Gaussians with constant-density ellipsoids and deriving the exact volumetric rendering equation along each camera ray. While effective, this approach reduces rendering speed by more than an order of magnitude compared to standard 3D GS and does not provide controllable boundary sharpness, which is suboptimal when the input view is a soft blending of colors.


\section{Method}
\label{sec:method}

In this section, we first revisit the non-overlapping assumption of 3D Gaussian Splatting, discussing why it is a critical prerequisite for the efficiency of the standard 3D GS algorithm. Building on this analysis, we introduce our modifications to the 3D GS framework to approximate the effects of applying softmax competition to overlapping 3D Gaussians. 

\subsection{The non-overlapping assumption}

Following~\citep{ewa}, we start with the original volume rendering equation along a camera ray. Let $\mathbf{x}=(x,y)$ denote a pixel position on the image plane, which also specifies the ray passing through that pixel, and let $I(\mathbf{x})$ represent the light intensity reaching the pixel. Let $o(\mathbf{x},l)$ denote the extinction function, defining the rate of light occlusion at distance $l$ from the camera, and let $c(\mathbf{x},l)$ be the emission coefficient at distance $l$. Then the original volume rendering equation is 
\begin{equation}
    I(\mathbf{x}) = \int_0^L c(\mathbf{x},l)o(\mathbf{x},l)T(o,\mathbf{x},l)dl
    \label{eq:vol_rend}
\end{equation}
where $T(o,\mathbf{x},l)=e^{-\int_0^l o(\mathbf{x},\mu)d\mu}$ denotes the transmittance until distance $l$. Now, suppose $o$ is the summation of $K$ kernel functions
\begin{equation}
    o(\mathbf{x},l) = \sum_{k=1}^K o_k(\mathbf{x},l).\label{eq:decom}
\end{equation}
 Within the context of 3D GS, each $o_k$ is represented as a 1D-projection of a 3D Gaussian to the camera ray, associated with a constant color $c_k$, where each Gaussian is limited to a compact support (e.g., $3\sigma$). Now, 3D GS further makes the key assumption that \textit{the support of all the trimmed Gaussian kernels do not overlap}, so that they can be sorted from nearest to farthest, indexed from $1$ to $K$. These assumptions allowed 
rewriting $T(o, \mathbf{x}, l)$ as 
the contribution of the first $k_l-1$ Gaussians that are closer to the camera than the distance $l$, and the contributions of the Gaussians can be viewed as independent to $l$ due to their non-overlapping support. In this manner, we can extract $T(o, \mathbf{x}, l)$ outside of the integral in Eq.~(\ref{eq:vol_rend}), leading to a greatly simplified formulation commonly utilized  in the literature.

We believe it is important to reproduce the original derivation of 3D GS (from the 1990s)~\cite{ewa} to highlight the assumptions that were used. Note we also ignores self-occlusion (of the $k_l$-th Gaussian which overlaps with $l$), and utilizes a Taylor expansion $e^x \approx 1 + x$ below:
\begin{align}
    T(o,\mathbf{x},l)&=e^{-\int_0^l o(\mathbf{x},\mu)d\mu}
    \approx e^{-\int_0^l \left(\sum_{j=1}^{k_l-1} o_j(\mathbf{x},\mu)\right)d\mu} \nonumber\\
    \tag*{\parbox[t]{.9\linewidth}{\raggedleft non-overlapping assumption, ignore self-occlusion}}\\
    &\approx e^{-\int_0^L \left(\sum_{j=1}^{k_l-1} o_j(\mathbf{x},\mu)\right)d\mu} \nonumber \\
    \tag*{\parbox[t]{.7\linewidth}{\raggedleft non-overlapping assumption}} \nonumber \\ &= e^{\sum_{j=1}^{k_l-1}\left(-\int_0^L  o_j(\mathbf{x},\mu)d\mu\right)} 
    = \prod_{j=1}^{k_l-1} e^{-\int_0^L  o_j(\mathbf{x},\mu)d\mu} \nonumber\\
    &\approx \prod_{j=1}^{k_l-1} \left(1-\int_0^L  o_j(\mathbf{x},\mu)d\mu\right). \label{eq:tapprox}\\
    \tag*{\parbox[t]{.7\linewidth}{\raggedleft Taylor expansion}}
\end{align}
where $l$ is changed to $L$ because the non-overlapping assumption dictates that the support of each $o_j(\mathbf{x},\mu)$ ends before the distance $l$ (hence $o_j(\mathbf{x}, \mu) = 0$ for $\mu  > l$). Importantly, $T(\cdot)$ is now only a piecewise-constant function of $l$, and its discontinuities coincide with the support of each $o_j$, hence the volume rendering equation becomes:
\begin{align}
    I(\mathbf{x}) &= \int_0^L \left(\sum_{k=1}^K c_k o_k(\mathbf{x},l)\right)T(o,\mathbf{x},l)dl\label{eq:orig} \\
    & \approx \sum_{k=1}^K c_k \left(\int_0^L o_k(\mathbf{x},l)dl\right) \prod_{j=1}^{k-1} \left(1-\int_0^L o_j(\mathbf{x},\mu)d\mu\right)\nonumber
\end{align}
where we have successfully switched the order of integration and summation utilizing the piecewise-constant $T(\cdot)$ approximation in Eq.~(\ref{eq:tapprox}). Integrating each 1D Gaussian along the ray (denoted by $a_k = \int_0^L o_k(\mathbf{x},l) dl$), we arrive at the commonly used  Gaussian splatting formulation
\begin{equation}
    I(\mathbf{x}) = \sum_{k=1}^K c_k a_k \prod_{j=1}^{k-1} \left(1-a_j\right).
    \label{eq:gs}
\end{equation}
From this derivation, it is clear that the non-overlapping assumption is key for the efficiency of 3D GS: it allows the 3D Gaussians to be integrated along the ray first, reducing the primitives to 2D, after which they can be composed efficiently using the standard splatting equation—i.e., ``splat first, compose second."

Although the non-overlapping assumption simplifies the integration process, it is inherently unrealistic. In real-world scenes represented with 3D Gaussians, different colors are often adjacent, and Gaussians frequently lie in close proximity at color boundaries. Under the vanilla 3D GS framework, overlapping Gaussians are assigned an arbitrary fixed order and rendered according to this order-dependent equation. This leads to view inconsistency, since the ordering  changes when the viewpoint is rotated, leading to sudden, significant  difference in the rendering. To address these challenges, we propose \methodname{}, a unified approach that relaxes the non-overlapping assumption and enforces softmax-based competition among overlapping Gaussians.

\subsection{Softmax competition and approximation}\label{sec:approx}
\begin{figure}[h]
    \centering
    \includegraphics[width=0.9\linewidth]{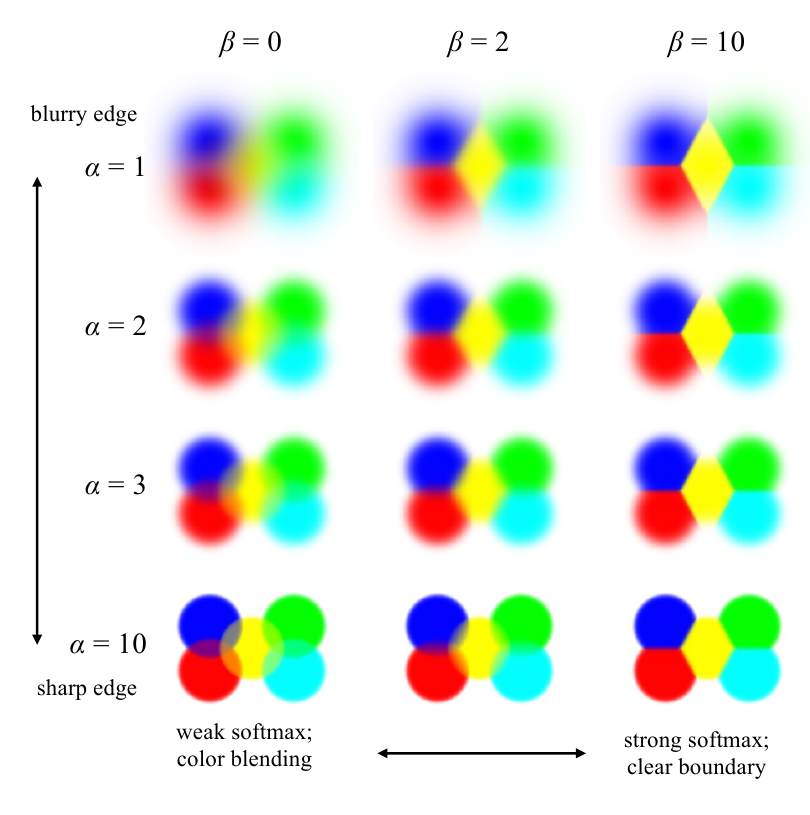}
    \vskip -0.2in
    \caption{\small We demonstrate the effect of parameter $\alpha$ and $\beta$ with five overlapping Gaussians. Parameter $\alpha$ controls the sharpness of individual Gaussian boundary and $\beta$ controls the harshness of the softmax competition between Gaussians, realizing smooth transition from color blending to clear boundary.}
    \vspace{-0.2in}
    \label{fig:smgs_param}
\end{figure}

Due to the inherently diffuse boundaries of Gaussians, representing the sharp color transitions common in real-world scenes requires stacking many tiny Gaussians along each edge. Simply relaxing the non-overlap assumption and allowing equal color blending between Gaussians does not resolve this blurriness issue (top-left of Fig.~\ref{fig:smgs_param}). Methods like GEF can sharpen a single Gaussian’s boundary but cannot control how two overlapping Gaussians should behave—whether they should blend smoothly or produce a crisp separation.

Ideally, two overlapping Gaussians should be able to adaptively transition between a winner-take-all behavior—assigning each pixel to a single dominant Gaussian for sharp boundaries—and a more balanced blending when appropriate. To achieve this, we introduce a softmax-based competition between overlapping Gaussians, controlled by a tunable parameter $\beta$ that modulates the strength of competition. This formulation enables a continuous spectrum of behaviors, ranging from smooth color averaging to well-defined boundaries.
Specifically, we re-formulate Eq.~(\ref{eq:decom}) as:
\begin{equation}
    o(\mathbf{x},l) = \sum_{k=1}^K w_k (\mathbf{x}, l,p) o_k(\mathbf{x},l)\label{eq:decom2}
\end{equation}
where $w_k$ denotes the softmax weight resulting from the competition among the Gaussian exponents $p_k$:
\begin{equation}
    w_k(\mathbf{x},l,p) = \dfrac{\exp(\beta\cdot p_k(\mathbf{x},l))}{\sum_{j=1}^{K}\exp(\beta\cdot  p_j(\mathbf{x},l))}.
    \label{eq:softmax_weight}
\end{equation}
Here, $o_k \sim \exp(p_k)$, and $\beta$ controls the strength of the softmax competition. When $\beta = 0$, Gaussian colors are blended equally, whereas a large $\beta$ produces sharp boundaries between overlapping Gaussians (see Fig.~\ref{fig:smgs_param}). We operate on $p_k$ instead of $o_k$ because $o_k$ ranges only from 0 to $\infty$; using $\exp(o_k)$ would yield values no less than 1, making it difficult for the softmax weights to diminish and producing undesired visual effects. We purposefully keep separate notations for $p_k$ and $o_k$ because as we later add individual Gaussian edge sharpness control in Sec.~\ref{sec:smgs_alg}, $o_k \sim \exp(p_k)$ will no longer hold.  Substituting Eq.~(\ref{eq:decom2}) into Eq.~(\ref{eq:orig}) gives
\begin{equation}
    I(\mathbf{x}) = \int_0^L \left(\sum_{k=1}^K c_k w_k(\mathbf{x},l,p) o_k(\mathbf{x},l)\right)T(o,\mathbf{x},l)dl\label{eq:vr_sm}
\end{equation}
. However, directly evaluating Eq.~(\ref{eq:vr_sm}) requires a full numerical integration, which is impractical for real-time rendering. Therefore, here we impose a mild assumption that the Gaussians can still be \textit{approximately} sorted along the ray—such that once a Gaussian loses dominance in the softmax competition, it does not regain it in latter parts of the ray. Through this more realistic assumption, we can derive an efficient approximation. Specifically, we first compute the per-Gaussian softmax-weighted integral $\mathring{a}_k = \int w_k(\mathbf{x},l,p) o_k(\mathbf{x},l)dl$ and then apply the standard splatting formulation
\begin{equation}
I(\mathbf{x}) = \sum_{k=1}^K c_k \mathring{a}_k \prod_{j=1}^{k-1} (1 - \mathring{a}_j),
\end{equation}
retaining the strategy of splatting first and compositing afterwards. 
This approximation is most accurate when $\beta$ is large (the winner-take-all regime), where Gaussians effectively behave as non-overlapping after softmax suppression. 

\begin{figure*}[h]
    \centering
    \includegraphics[width=\linewidth]{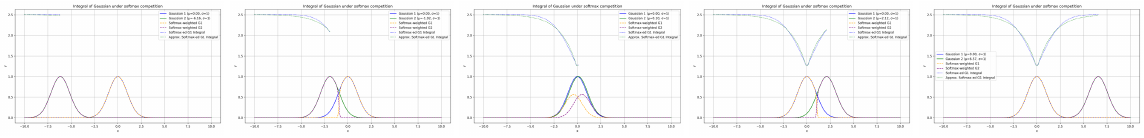}
    \vskip -0.05in
    \caption{\small 
    Visualization of softmax competition between two identical Gaussians ($o_j$, $o_k$). We plot their softmax-ed values ($\hat{o}_j, \hat{o}_k$) and the integral $\mathring{a}_k$ of $\hat{o}_k$ (blue dotted line). The influence of $o_j$ on $\mathring{a}_k$ decays nearly exponentially with distance, allowing $\mathring{a}_k$ to be approximated via an exponentially weighted linear interpolation between the $\mathring{a}_k$ values in the non-overlapping and fully overlapping cases (green dotted line).
}
    \label{fig:smgs_int}
\end{figure*}

However, the weight $w_k$ still depends on all other Gaussians through the softmax, making the exact computation of $\mathring{a}_k$ difficult, even though the non-softmax-ed integrals $a_k = \int_0^L o_k(\mathbf{x},l) dl$ are readily available. We therefore propose to apply the softmax competition directly on the set of $a_k$'s to obtain $\hat{a}_k$.
Consider the simple case where two Gaussians share the same shape and completely overlap, then $\mathring{a}_k=\hat{a}_k$; if they are fully separated, then $\mathring{a}_k=a_k$. Motivated by these two extremes, we approximate the full $\mathring{a}_k$ by interpolating between $a_k$ and $\hat{a}_k$.
Fig.~\ref{fig:smgs_int} plots $\mathring{a}_k$ of a Gaussian $o_k$ under the influence of another identical Gaussian $o_j$ 
as a function of the distance $|\mu_k - \mu_j|$ between them. Note as they separate, $\mathring{a}_k$ decays back to $a_k$ approximately exponentially. Therefore, we propose to use an exponentially decaying factor $s$ to interpolate between $a_k$ and $\hat{a}_k$.
Specifically, consider two 3D Gaussians that project to 2D splats with absorbance values $a_j, a_k$ and Gaussian exponents $p_j, p_k$ with $a_{\{j,k\}}\sim \exp(p_{\{j,k\}})$ at pixel $\mathbf{x}$. We compute
\begin{align}
    &\hat{a}_k = w_k a_k \ \ \ \  w_k=\tfrac{\exp(\beta p_k)}{\exp(\beta p_j)+\exp(\beta p_k)}\label{eq:sm}\\
    &s = \exp(-\gamma|d_k-d_j|)\label{eq:decay}\\
    &\bar{a}_k = s \hat{a}_k + (1-s)a_k\ \ \ \ 
    \bar{a}_j = s \hat{a}_j + (1-s)a_j\label{eq:interp}
\end{align}
where $d_{\{j,k\}}$ denote the depths of the splats from the camera, and $s$ is a $\gamma$-controlled decay factor that decreases with their depth difference. This formulation empirically captures how $\mathring{a}_k$ varies with the separation between two splats. Now we can present the central idea of \methodname{}: \textit{when two splats are extremely close, they are treated as belonging to the same surface and their colors are blended through softmax competition; as they separate, they are rapidly interpreted as distinct surfaces again}.

While the set of approximation equations above describe the two-Gaussian case, in practice a camera ray may intersect many Gaussians. This leads to our full \methodname{} algorithm, which generalizes the above approximation to an arbitrary number of overlapping splats.

\subsection{\methodname{} algorithm}\label{sec:smgs_alg}
In this section, we present the full \methodname{} algorithm. The method preserves the linear complexity of the rendering equation by processing Gaussians along each camera ray sequentially, from nearest to farthest. At the $k$-th Gaussian, we treat all previously processed Gaussians as a single entity and compute the softmax competition between this combined set and the current Gaussian. Specifically, let $T_{k-1} = \prod_{j=1}^{k-1} (1 - a_j)$ denote the current transmittance, and define $a_{\text{past}} = 1 - T_{k-1}$ as the accumulated absorbance. Let $d_{\text{past}}$ and $p_{\text{past}}$ be the moving average depth and Gaussian exponent of the past Gaussians, respectively, and let $p_{k}, a_{k}, d_{k}$ represent the exponent, absorbance, and depth of the current Gaussian. Applying the same set of equations (Eq.~(\ref{eq:sm})–(\ref{eq:interp})) then yields the updated values $\bar{a}_{\text{past}}$ and $\bar{a}_{k}$.

\textbf{Order invariance and transmittance maintenance.} However, two challenges remain. First, although $a_{\text{past}}$ and $a_{\text{k}}$ appear symmetric in the computations above, in the 3D GS algorithm the current Gaussian color is multiplied by $(1-\bar{a}_{\text{past}})\bar{a}_{k}$ rather than just by $a_{k}$, which reduces the contribution of the current Gaussian. Consider two Gaussian splats lying at the same depth. The one processed second according to the sorting order will have its color discounted by a factor of $1-\hat{a}_{\text{past}}$ relative to the first. The second issue is that both $\hat{a}_{\text{past}}<a_{\text{past}}$ and $\hat{a}_{k}<a_{k}$, since the softmax weights are less than one. This causes the output transmittance after the two Gaussians to be higher than in the scenario without softmax competition, causing transmittance inconsistency between image regions with and without overlap. To address these issues, we modify $\hat{a}_{\text{past}}$ and $\hat{a}_{k}$ to ensure order invariance and maintain the original output transmittance. Formally, setting $T_{k}=\prod_{j=1}^k (1-a_j) = (1-a_{\text{past}})(1-{a}_{k})$, we aim to find $\tilde{a}_{\text{past}}$ and $\tilde{a}_{k}$ satisfying
\begin{align}
    \tfrac{(1-\tilde{a}_{\text{past}})\tilde{a}_{k}}{\tilde{a}_{\text{past}}}&=\tfrac{\hat{a}_{k}}{\hat{a}_{\text{past}}}\\
\tag*{\parbox[t]{.9\linewidth}{\raggedleft order invariance}}\\
    (1-\tilde{a}_{\text{past}})(1-\tilde{a}_{k}) &= T_{k}.\\
    \tag*{\parbox[t]{.9\linewidth}{\raggedleft maintain original transmittance}}
\end{align}
Solving this system of equations yields
\begin{align}
    \tilde{a}_{\text{past}}=\tfrac{\hat{a}_{\text{past}}(1-T_{k})}{\hat{a}_{\text{past}}+\hat{a}_{k}}, 
    \tilde{a}_{k}=\tfrac{\hat{a}_{k}(1-T_{k})}{\hat{a}_{k}+ \tilde{a}_{\text{past}}T_{k}}
\end{align}
We then apply Eq.~(\ref{eq:interp}) to $\tilde{a}_{\text{past}}$ and $\tilde{a}_{k}$ to obtain $\bar{a}_{\text{past}}$ and $\bar{a}_{k}$. After interpolation, however, the output transmittance may again deviate from $T_{k}$. We therefore further correct the absorbance values by computing a scaling factor $m$ such that
\begin{equation}
    (1-m \bar{a}_{\text{past}})(1-m \bar{a}_{k})=T_{k}.
    \label{eq:m_solution}
\end{equation}
which leads to the final approximation $\mathring{a}_{\text{past}} = m \bar{a}_{\text{past}}$ and $\mathring{a}_{k} = m \bar{a}_{k}$. Note that $m$ can be solved exactly from eq.~(\ref{eq:m_solution}), see the supplementary material.

\textbf{Single Gaussian boundary sharpness.} While softmax competition can enforce sharp boundaries between two or more Gaussians, it cannot sharpen an individual Gaussian by itself. To address this, we incorporate the Generalized Exponential Function (GEF) first adopted by GES~\citep{ges}, where the squared distance $\|\cdot\|^2$ in the Gaussian function is replaced by $\|\cdot\|^{2\alpha}$, allowing flexible control over the shape of Gaussian boundaries. Unlike GES, we implement the exact GEF directly in our CUDA kernel rather than using an approximate rasterization. Note that this modification affects only the absorbance value $a$, not the Gaussian exponent $p$, which participates in the softmax competition. Modifying $p$ directly would cause the softmax competition less effective at distinguishing between Gaussians as the boundary sharpness increases.

Collectively, we introduce three additional parameters for each Gaussian: $\alpha$, which controls the boundary sharpness of an individual Gaussian; $\beta$, which governs the strength of softmax competition between overlapping Gaussians; and $\gamma$, which determines the decay rate of the softmax competition with increasing distance between Gaussians. 
The complete forward pass of \methodname{} rendering algorithm is provided in supplementary material.

\textbf{Efficient backward pass.}
The backward pass of the original 3D GS algorithm also enjoys linear complexity, proceeding from the farthest Gaussian to the nearest along each ray and computing gradients in reverse. This is achieved by inferring the input transmittance $T_{\text{in}}$ from the output transmittance $T_{\text{out}}$ and the current absorbance $a_{\text{cur}}$:
\begin{equation}
    T_{\text{in}}=\frac{T_{\text{out}}}{1-a_{\text{cur}}}.
\end{equation}
However, in \methodname{} this inference is no longer valid, and many other intermediate values necessary for gradient computation ( e.g. $a_{\text{past}}$ and $d_{\text{past}}$) cannot be directly obtained from the output values at each Gaussian iteration. To preserve the linear complexity of the backward pass, we  apply \methodname{} only to the closest $K$ Gaussians for each image patch. This allows us to create CUDA arrays of length $K$ and cache all intermediate values required for gradient computation during a single forward pass. During the backward pass, the $k$-th entry of these arrays can be directly accessed, eliminating the need to recompute the forward pass up to the $k$-th Gaussian. The hyperparameter $K$ can be empirically chosen based on the approximate total number of Gaussians in the scene. We use $K=128$ for all the real-world scene experiments, which  covered $70\%$ - $93\%$ of the Gaussians.


\section{Experiments}
\label{sec:experiment}

\begin{figure}[h]
    \centering
    \includegraphics[width=\linewidth]{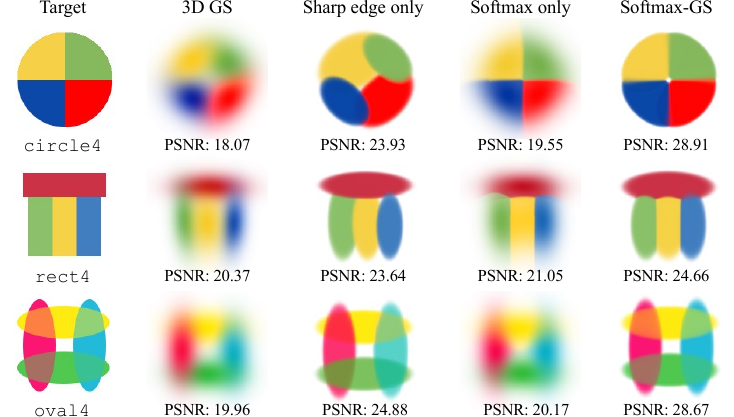}
    \vskip -0.05in
    \caption{\small Simple geometry fitting with 4 Gaussians using 3D GS, \methodname{}, and two variants ablating softmax competition or boundary sharpness. Only one component yields suboptimal results, while \methodname{} achieves the best of both. Optimization process videos are available on the project page.} 
    \label{fig:smgs_simple}
\end{figure}

\begin{figure}[h]
    \centering
    \includegraphics[width=\linewidth]{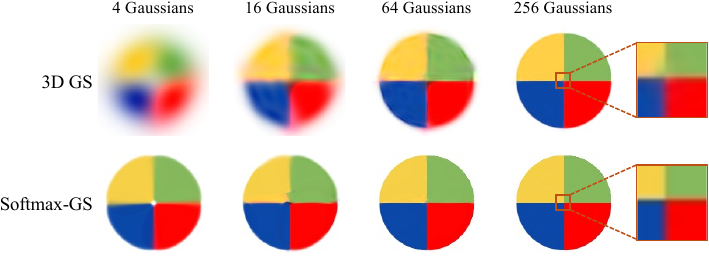}
    \vskip -0.05in
    \caption{\small Fitting experiment with 3D GS and \methodname{} on \texttt{circle4} with different Gaussian number constraint. As the number of Gaussians increases, 3D GS achieves similar result as \methodname{}, but still struggles with sharp corners.} 
    \label{fig:smgs_simple2}
\end{figure}

\setlength{\tabcolsep}{3pt}

\begin{table*}[h]
\centering

\begin{minipage}[t]{0.8\linewidth}
\begin{center}
    \captionsetup{type=table}
    \hspace{-0.1in}\resizebox{\linewidth}{!}{
\begin{scriptsize}
\renewcommand{\arraystretch}{1.2}
\begin{tabular}{@{}$l@{\ \ }^c@{\ \ }^c@{\ \ }^c@{\ \ }^c@{\ \ }^c@{\ \ }^c@{\ \ \ \ }^c@{\ \ }^c@{\ \ }^c@{\ \ }^c@{\ \ }^c@{\ \ }^c@{\ \ \ \ }^c@{\ \ }^c@{\ \ }^c@{\ \ }^c@{\ \ }^c@{\ \ }^c@{\ \ }^c@{}}
\toprule
\textbf{Pattern} &\multicolumn{6}{^c}{\texttt{circle4}} & \multicolumn{6}{^c}{\texttt{rect4}} & \multicolumn{6}{^c}{\texttt{oval4}}\\
 \cmidrule(lr){2-7}\cmidrule(lr){8-13}\cmidrule(lr){14-19}
\textbf{Method\textbackslash \#GS} &4 & 16 & 64 & 256 & 1024 & 4096 &  4 & 16 & 64 & 256 & 1024 & 4096 & 4 & 16 & 64 & 256 & 1024 & 4096\\
 \midrule
 3D GS &18.07 & 22.16 & 28.16 & 36.47 & 45.00 & 44.98 & 20.37 & 23.79 & 30.51 & 41.54 & 53.04 & 59.91 & 19.96 & 22.28 & 27.61 & 35.61 & 44.69 & 49.21 \\
 Sharp edge only &23.93 & 30.91 & 42.11 & 48.17 & 48.53 & 49.91 & 23.64 & 29.32 & 46.72 & 53.76 & 57.24  & 59.38 & 24.88 & 28.77 & 38.49 & 46.48 & 48.10 & 50.22 \\
Softmax only &19.55 & 23.57 & 29.64 & 38.40 & 47.38  & 48.98 & 21.05 & 24.98 & 30.28 & 44.08 & 53.58  & 59.03 & 20.17 & 22.34 & 27.50 & 36.43 & 44.39 & 49.65 \\
\methodname{} &\cellcolor{first}28.91 & \cellcolor{first}32.51 & \cellcolor{first}42.27 & \cellcolor{first}49.14 & \cellcolor{first}50.18 & \cellcolor{first}51.93  &
\cellcolor{first}24.66 & \cellcolor{first}30.29 & \cellcolor{first}50.16 & \cellcolor{first}54.97 & \cellcolor{first}57.95 &  \cellcolor{first}60.06 &
\cellcolor{first} 32.51 & \cellcolor{first}33.82 & \cellcolor{first}41.42 & \cellcolor{first}46.72 & \cellcolor{first}48.70  & \cellcolor{first}50.37 \\
\bottomrule
\end{tabular}
\end{scriptsize}}
\end{center}
\vspace{-0.2in}
\caption{Quantitative results (PSNR) on simple geometry fitting experiments. All methods are optimized for 10K steps with a limit on the maximum number of Gaussians.}
\label{tb:smgs_simple}
\end{minipage}
\vskip -0.1in
\end{table*}

\subsection{Simple geometry patterns}

To clearly demonstrate the functionality of \methodname{}, we created several simple color patterns as shown in Fig.~\ref{fig:smgs_simple}. We initialize 4 black Gaussians and run optimization for 10K steps without opacity reset using default rendering losses. 

In Fig.~\ref{fig:smgs_simple}, we show optimization results with only 4 Gaussians. \methodname{} achieves results closest to the target, highlighting the complementary roles of edge sharpness control and softmax competition. Using edge sharpness alone improves PSNR significantly, but the Gaussians on the same surface are still enforced with an order, making color blending or correct boundary between Gaussians difficult to obtain. Conversely, using only softmax competition struggles to control the edge sharpness of the outside boundaries of the individual Gaussians. 

\setlength{\tabcolsep}{3pt}

\begin{table*}[htb]
\centering
\begin{minipage}[t]{0.9\linewidth}
\begin{center}
    \captionsetup{type=table}
    \hspace{-0.1in}\resizebox{\linewidth}{!}{
\begin{scriptsize}
\renewcommand{\arraystretch}{1.2}
\begin{tabular}{@{}$l^c^c^c^c^c^c^c^c^c^c^c^c@{}}
\toprule
\rowstyle{\bfseries}
 \multirow{2}{*}{Method} &  \multicolumn{4}{^c}{Mip-NeRF360} & \multicolumn{4}{^c}{Tanks\&Temples} & \multicolumn{4}{^c}{Deep Blending} \\
 \cmidrule(lr){2-5}\cmidrule(lr){6-9}\cmidrule(lr){10-13}
 & PSNR$\uparrow$ & SSIM$\uparrow$ & LPIPS$\downarrow$ & \#GS $\downarrow$ & PSNR$\uparrow$ & SSIM$\uparrow$ & LPIPS$\downarrow$ & \#GS$\downarrow$& PSNR$\uparrow$ & SSIM$\uparrow$ & LPIPS$\downarrow$ & \#GS$\downarrow$ \\
 \midrule
  3D GS$_{5\%}$ & 25.05 & 0.744 & 0.345 & 104K & 20.5 & 0.746 & 0.330 & 112K & 27.53 & 0.854 & 0.355 & 82K \\
 \textbf{\methodname{}$_{5\%}$} & \cellcolor{first}26.65 & \cellcolor{first}0.769 & \cellcolor{first}0.328 & 105K & \cellcolor{first}22.0 & \cellcolor{first}0.777 & \cellcolor{first}0.308 & 113K & \cellcolor{first}28.74 & \cellcolor{first}0.872 & \cellcolor{first}0.339 & 85K \\
 \midrule
GES~\cite{ges} &  28.64 & \cellcolor{first}0.855 & \cellcolor{second} 0.214 & 1407K &  23.24 & \cellcolor{second}0.824 & 0.238 & 947K &  29.43 & \cellcolor{second} 0.889 & 0.287 & 1541K \\
3DCS~\cite{convexgs} &  \cellcolor{second}28.86 & \cellcolor{second} 0.849 & \cellcolor{first} 0.195 & 1139K &  23.25 & 0.819 & \cellcolor{second} 0.233 & 443K &  29.24 & 0.885 & \cellcolor{first} 0.268 & 1375K \\
\textbf{\methodname{}$_{\text{mini}}$} & 28.16 & 0.817 & 0.261 & 548K & \cellcolor{second} 23.51 & 0.813 & 0.257 & 520K & \cellcolor{second}29.59 & 0.887 & 0.293 & 777K \\
\textbf{\methodname{}$_{\text{light}}$} &  \cellcolor{first}29.05 & \cellcolor{first}0.856 & \cellcolor{second}0.216 & 1279K & \cellcolor{first}23.93 & \cellcolor{first} 0.833 & \cellcolor{first} \cellcolor{first} 0.229 & 1052K & \cellcolor{first} 29.66 & \cellcolor{first} 0.892 & \cellcolor{second} 0.282 & 1440K \\
\midrule
3D GS~\cite{3dgs} &  29.16 & 0.869 & 0.185 & 2593K &  23.66 & 0.832 & 0.215 & 1900K &  29.45 & 0.882 & 0.277 & 2536K \\
3D-HGS~\cite{3dhgs} & 29.61 & 0.871 & 0.178 & 2874K &  24.17 & 0.843 & 0.206 & 1940K & 28.88 & 0.882 & 0.286 & 2872K \\
StopThePop~\cite{stopthepop} &  28.84 & 0.871 & 0.180 & 3070K &  23.01 & 0.831 & 0.215 & 1929K & 29.49 & \cellcolor{first}0.891 & \cellcolor{first}0.272 & 2965K \\
EVER\cite{ever} & 29.06 & 0.871 & 0.177 &  3058K & 23.32 & 0.850 & \cellcolor{first}0.183 & 5124K & 29.09 & 0.890 & 0.277 & 3382K \\ 
LC-WSR~\cite{sortfree} & 29.03 & 0.858 & 0.189 & 3648K & 22.90  &  0.819 & 0.223 & 4069K &  29.48 & 0.888 & 0.276 & 3230K \\

GS-MCMC~\cite{gsmcmc} & \cellcolor{second}29.62 & \cellcolor{first}0.888 & \cellcolor{first}0.161 & 2593K & \cellcolor{second}24.40 & \cellcolor{first}0.853 & 0.189 & 1900K &  29.35 & 0.890 & 0.276 & 2536K \\
SSS~\cite{sss} & \cellcolor{second}29.63 & 0.885 & \cellcolor{first}0.159 & 2593K & \cellcolor{second}24.39 & \cellcolor{first}0.854 & \cellcolor{second}0.186 & 1900K & 28.82 & 0.887 & 0.275 & 2536K \\
\textbf{\methodname{}} &  29.41 & 0.871 & 0.184 & 2495K & 24.09 & 0.840 & 0.211 & 1873K & \cellcolor{first}29.77 & \cellcolor{first}0.892 & 0.276 & 2402K \\
\textbf{\methodname{}-MCMC} & \cellcolor{first}29.73 & \cellcolor{first}0.888 & \cellcolor{first}0.160 & 2593K & \cellcolor{first}24.52 & \cellcolor{first}0.854 & 0.189 & 1900K &  \cellcolor{second}29.68 & \cellcolor{first}0.893 & \cellcolor{first}0.273 & 2536K \\
\bottomrule
\end{tabular}
\end{scriptsize}}
\end{center}
\vspace{-0.5cm}
\caption{\footnotesize Image rendering quality comparison with state-of-the-art methods on real-world datasets, with all methods optimized for 30K steps. The first group includes approaches using sparse Gaussian representations, while the second group compares methods with a similar number of Gaussians as the standard 3D GS setup. Best results within each group are highlighted in \textcolor{first}{red}, and second-best results in \textcolor{second}{orange}.
}
\label{tb:smgs_real}
\end{minipage}
\vskip -0.15in
\end{table*}

\setlength{\tabcolsep}{1pt}
\begin{table}[htb]
\centering
\begin{minipage}[t]{0.9\linewidth}
\begin{center}
    \captionsetup{type=table}
    \hspace{-0.1in}\resizebox{\linewidth}{!}{
\begin{scriptsize}
\renewcommand{\arraystretch}{1.2}
\begin{tabular}{@{}$l^c^c^c^c@{}}
\toprule
\rowstyle{\bfseries}
 \multirow{2}{*}{Method} &  \multicolumn{2}{^c}{Training time (min)$\downarrow$} & \multicolumn{2}{^c}{Rendering speed (FPS)$\uparrow$} \\
 \cmidrule(lr){2-3}\cmidrule(lr){4-5}
\rowstyle{\ttfamily} & bicycle & train & bicycle & train\\
 \midrule
3D GS & \cellcolor{second}36 & \cellcolor{first}27 & 54 & \cellcolor{second} 98\\
StopThePop & 37 & 28 & 47 & 49 \\
3DCS  & 52 & 58 & 42 & 36 \\
EVER & 77 & 56 & 8 & 12 \\
\methodname{}$_{\text{mini}}$ & \cellcolor{first}26 & \cellcolor{first}27 & \cellcolor{first}152 & \cellcolor{first} 166 \\
\methodname{}$_{\text{light}}$ & 37 & 32 & \cellcolor{second}64 & 95 \\
\methodname{} & 45 & 32 & 44 & 77 \\
\bottomrule
\end{tabular}
\end{scriptsize}}
\end{center}
\vspace{-0.5cm}
\caption{\footnotesize Training and rendering speed comparison.
}
\label{tb:smgs_profile}
\end{minipage}
\vskip -0.2in
\end{table}

We also conduct experiments varying the maximum number of Gaussians allowed during the densification process. Quantitative results in terms of PSNR are provided in Table~\ref{tb:smgs_simple}. As expected, the gap between vanilla 3D GS and \methodname{} decreases when more Gaussians are used, since 3D GS can approximate sharp boundaries by stacking many small Gaussians. Nevertheless, \methodname{} maintains an advantage w.r.t. 3D GS and \textit{sharp edge only} even with 1024 Gaussians. Qualitative comparisons are shown in Fig.~\ref{fig:smgs_simple2}.

\subsection{Real-world data}

We evaluate \methodname{} on a standard real-world benchmark suite, including seven scenes from Mip-NeRF360~\citep{mip360}, two scenes from Tanks\&Temples~\citep{tanks} (\texttt{train} and \texttt{truck}), and two scenes from DeepBlending~\citep{deepblending} (\texttt{drjohnson} and \texttt{playroom}). For DeepBlending, we use the original image resolution. For Mip-NeRF360 and Tanks\&Temples, we preserve the aspect ratio while resizing the width to 1600.

For real-world scenes, 
we adopt the same densification and opacity resetting strategy as in the original 3D GS, while raising the Gaussian size pruning threshold to 0.4, as the GEF-based edge sharpening might shrink the effective size of the Gaussian splats, and our strategy allows the usage of larger and \textbf{fewer} Gaussians to represent color patterns. We also present the \methodname{}$_{5\%}$, \methodname{}$_\text{mini}$  and \methodname{}$_\text{light}$ versions with significantly fewer Gaussians by tuning the densification parameter to 4\textsc{e}-3, 5\textsc{e}-4 and 3\textsc{e}-4, respectively. 
In addition to running \methodname{} with the standard 3D GS optimization, we also evaluate its performance combined with GS-MCMC’s improved optimization strategy. Notably, our transmittance-maintaining rendering algorithm integrates seamlessly with GS-MCMC’s cloning approach, as it ensures that overlapping “new” Gaussians conform to the overall transmittance assigned to them.

\begin{figure*}[htb]
    \centering
    \includegraphics[width=0.81\linewidth]{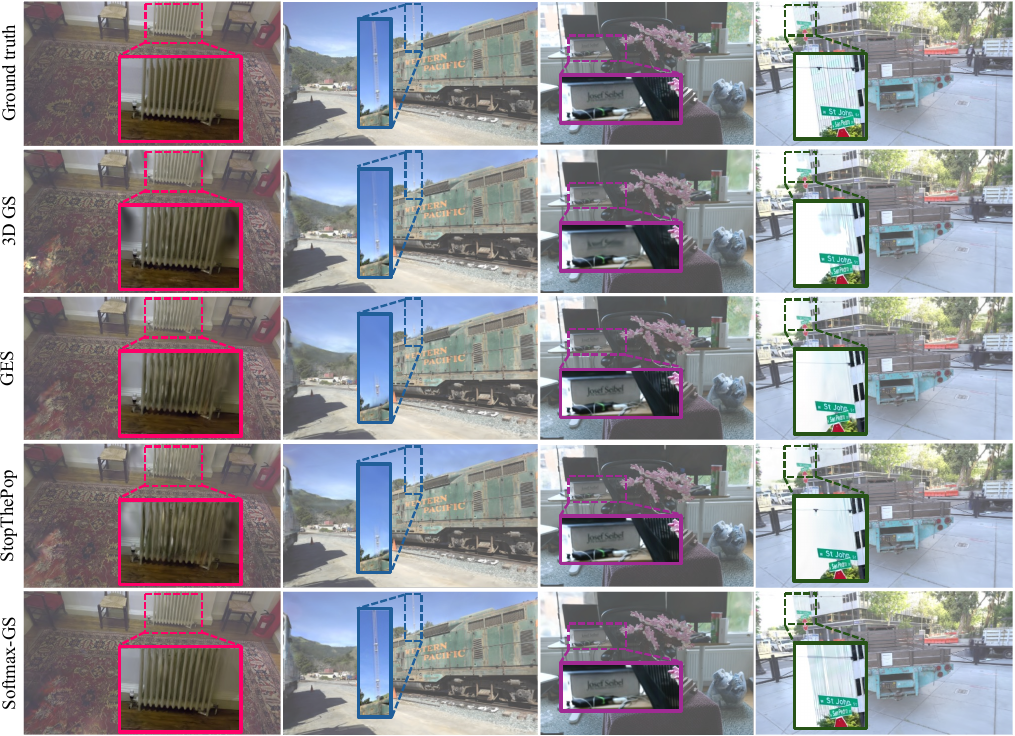}
    \vskip -0.05in
    \caption{\small Qualitative comparison with 3D GS, GES and StopThePop on real world datasets.} 
    \label{fig:smgs_rw}
    \vskip -0.15in
\end{figure*}

Quantitative comparisons are shown in Table~\ref{tb:smgs_real} in  three categories: (1) methods evaluated under extreme sparsity constraints, (2) methods designed to produce sparse scene representations~\cite{ges,convexgs}, and (3) methods using a number of Gaussians comparable to standard 3D GS. In the first group, under a stringent budget of only 5\% of the original Gaussians, \methodname{} consistently outperforms 3D GS by $>$1 dB PSNR across all datasets, demonstrating stronger robustness to extreme sparsity. In the second group, \methodname{}$_{\text{light}}$ substantially outperforms all competing sparse approaches—particularly in PSNR—demonstrating its ability to achieve higher fidelity with fewer Gaussians. Remarkably, \methodname{}$_{\text{light}}$ attains performance on par with the original 3D GS while using only about \textbf{half} as many Gaussians.
In the third group, both \methodname{} and \methodname{}-MCMC consistently outperform 3D GS and GS-MCMC, respectively. In particular, \methodname{} improves PSNR by approximately +0.3 dB across all datasets compared to 3D GS. 

The qualitative comparison between \methodname{} and the baselines is provided in Fig.~\ref{fig:smgs_rw}. \methodname{} achieves significantly better fidelity on thin structures such as (from left to right) the radiator, the antenna, the text and the details of the building. Such details are important for faithful reconstruction of the scene.

\subsection{Training and rendering speed}
Table \ref{tb:smgs_profile} presents the training and rendering speed in two representative scenes, \texttt{bicycle} and \texttt{train}. 
Our method preserves linear time complexity with respect to the number of Gaussians per ray for both forward and backward passes. The additional softmax-competition and transmittance-maintenance steps add only about 1.2× increase in training time, while rendering runs at roughly 80\% of the original 3D GS speed. Furthermore,  \methodname{}$_\text{light}$ matches the training and rendering speeds of standard 3D GS due to the reduced Gaussian count.

\section{Conclusion}
\label{sec:conclusion}
We propose \methodname{}, a unified solution to the view-inconsistency and diffuse-boundary issues in the original 3D GS algorithm. Our method introduces a softmax-based color-merging mechanism for overlapping Gaussians with controllable competition strength, enabling a continuum of visual behaviors ranging from smooth color blending to sharp, winner-take-all boundaries. Softmax-GS is derived by carefully modifying the assumption from the original volume rendering equation and designed to ensure order invariance for overlapping Gaussians, while maintaining consistent transmittance across the overlapping and non-overlapping regions, thereby preventing discontinuity artifacts. 
\methodname{} achieves state-of-the-art performance on real-world benchmarks for both sparse and dense Gaussian reconstructions. Providing both boundary sharpness control and view consistency, \methodname{} offers a flexible and efficient framework for real-world scene reconstruction.
\section*{Acknowledgements}
Chen Ziwen and Li Fuxin are partially supported by Oregon State University College of Agricultural Sciences Seed grant AGD010-AS06, NSF grants 1751402 and 2321851, Oregon Department of Agriculture 2023 Specialty Crop Block Grant, DARPA TIAMAT grant HR0011-24-9-0423 and an Adobe Fellowship.
{
    \small
    \bibliographystyle{ieeenat_fullname}
    \bibliography{main}
}

\clearpage
\setcounter{page}{1}
\maketitlesupplementary

\section{More implementation details}

For simple-geometry fitting experiments, we place a camera at the origin facing the +z direction, and initialize  four black Gaussians with identical shapes slightly apart at a depth of 1 unit in front of the camera. Optimization is run for 10K steps without opacity reset against the target image using default rendering losses. We set the learning rates for $\alpha$, $\beta$, and $\gamma$ to $0.0003$, $0.0003$, and $0.0002$, respectively.
For real-world benchmarks, we set the learning rates for $\alpha$, $\beta$, and $\gamma$ to $0.0008$, $0.008$, and $0.0004$, respectively. To accelerate training, we adopt the tile culling strategies proposed in \cite{speedysplat, stopthepop}.

To improve stability, we further introduce a variance regularization on $\beta$ and $\gamma$ along each camera ray. This term penalizes abrupt changes in these parameters when the depth ordering of Gaussians changes. Specifically, the variance is computed over Gaussians intersecting a ray and weighted by the inverse distance to the nearest (front-most) Gaussian, so that the regularization primarily affects the first visible cluster. The weights for both variance terms are set to $\lambda = 0.01$.

For our quantitative comparison, since previous works may have used different image resolutions for their benchmarks, we download the official code of the baselines and re-run all algorithms at a consistent resolution. We keep all methods’ default hyperparameters unchanged, and all scenes are optimized for 30K steps. 
For GS-MCMC~\citep{gsmcmc}, SSS~\citep{sss}, and \methodname-MCMC, all of which employ Monte Carlo-based optimization, we initialize from the SfM point cloud and cap the maximum number of Gaussians to match that of the original 3D GS.

\setlength{\tabcolsep}{3pt}
\begin{table*}[htb]
\centering
\begin{minipage}[t]{0.9\linewidth}
\begin{center}
    \captionsetup{type=table}
    \hspace{-0.1in}\resizebox{\linewidth}{!}{
\begin{scriptsize}
\renewcommand{\arraystretch}{1.2}
\begin{tabular}{@{}$l^c^c^c^c^c^c^c^c^c^c^c^c^c^c^c@{}}
\toprule
\rowstyle{\bfseries}
 \multirow{2}{*}{Method} &  \multicolumn{3}{^c}{\texttt{bicycle}} & \multicolumn{3}{^c}{\texttt{counter}} & \multicolumn{3}{^c}{\texttt{kitchen}} & \multicolumn{3}{^c}{\texttt{train}} & \multicolumn{3}{^c}{\texttt{truck}}\\
 \cmidrule(lr){2-4}\cmidrule(lr){5-7}\cmidrule(lr){8-10}\cmidrule(lr){11-13}\cmidrule(lr){14-16}
 & PSNR$\uparrow$ & MSE$\downarrow$ & \flip$_7\downarrow$ & PSNR$\uparrow$ & MSE$\downarrow$ & \flip$_7\downarrow$ & PSNR$\uparrow$ & MSE$\downarrow$ & \flip$_7\downarrow$ & PSNR$\uparrow$ & MSE$\downarrow$ & \flip$_7\downarrow$  & PSNR$\uparrow$ & MSE$\downarrow$ & \flip$_7\downarrow$  \\
 \midrule
3D GS & 25.13 & 0.017 & 0.032 &  29.26 & 0.0037 &0.016 &  31.42 & \cellcolor{first}0.0045 & 0.019 &  22.19 & 0.0126 & 0.048 & 25.12 & 0.0106 & 0.031 \\
StopThePop &  25.14 & \cellcolor{first}0.016 & \cellcolor{first}0.028 &  28.91 & 0.0037 & \cellcolor{first}0.015 &  31.33 & 0.0047 & 0.019 &  21.45 & \cellcolor{first}0.0121 & \cellcolor{first}0.044  & 24.56 & \cellcolor{first}0.0105 & 0.030 \\
\methodname{} &  \cellcolor{first}25.24 & \cellcolor{first}0.016 & 0.031 &  \cellcolor{first}29.38 & 0.0037 & 0.016 & \cellcolor{first} 32.11 & \cellcolor{first}0.0045 & \cellcolor{first}0.018 & \cellcolor{first} 22.62 & 0.0124 & \cellcolor{first}0.044  & \cellcolor{first}25.55 & \cellcolor{first}0.0105 & \cellcolor{first}0.029 \\
\bottomrule
\end{tabular}
\end{scriptsize}}
\end{center}
\vspace{-0.5cm}
\caption{\footnotesize View consistency measurement on five representative scenes.
}
\label{tb:view_consist}
\end{minipage}
\end{table*}

\section{Measurement of view consistency}

We evaluate the view consistency of \methodname{} following the protocol from StopThePop~\cite{stopthepop}. We first render a video from the reconstructed Gaussians and, for each frame $\mathbf{I}_i$, pair it with the frame seven steps ahead, $\mathbf{I}_{i+7}$. We then apply the RAFT optical-flow method~\cite{raft} to warp $\mathbf{I}_i$ to $\mathbf{I}_{i+7}$, producing $\hat{\mathbf{I}}_i$, and compute MSE and \flip$_7$ similarity metrics between $\hat{\mathbf{I}}_i$ and $\mathbf{I}_{i+7}$. Because \cite{stopthepop} does not disclose the camera-trajectory generation process, we simply insert 128 interpolated frames between each pair of target frames. Following \cite{stopthepop}, we run RAFT using the model pre-trained on SINTEL~\cite{sintel} and resize the input frames to SINTEL’s resolution of 1024×436.

We present detailed evaluation results on five representative scenes in Table~\ref{tb:view_consist}, reporting both rendering quality (PSNR) and view-consistency metrics (MSE and \flip). All methods are evaluated under the same protocol. We observe that \methodname{}, by addressing overlapping Gaussians, and StopThePop, by introducing a re-sorting mechanism, each improves view consistency individually. However, as shown in the rendered videos provided the supplementary materials, \methodname{} exhibits significantly fewer ``floater" artifacts compared to StopThePop (especially in the \texttt{kitchen} scene), indicating that our approach leads to more stable optimization behavior.

\section{Depth rendering of \methodname{}}
\begin{figure}
    \begin{minipage}[t]{0.9\linewidth}
    \includegraphics[width=0.99\linewidth]{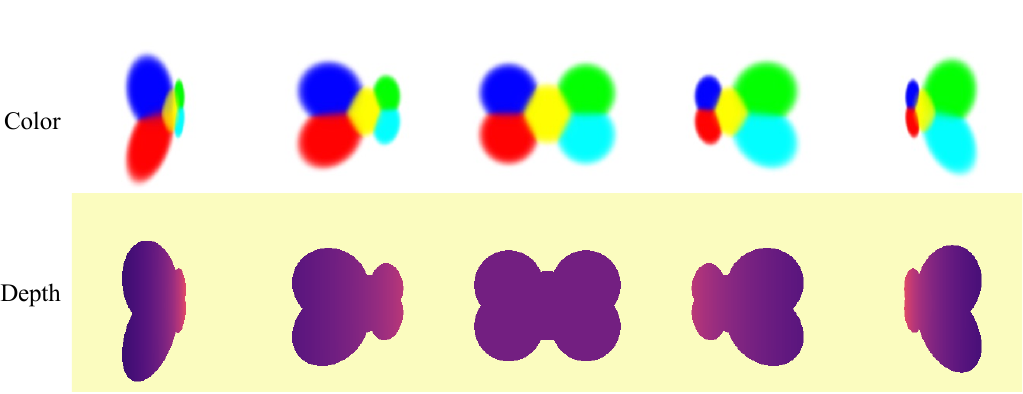}
    \caption{\footnotesize Pixel-wise depth rendering of synthetic pattern.}
    \label{fig:dep_synth}
    \end{minipage}
\vspace{-0.1in}
\end{figure}

\begin{figure}
    \centering
    \begin{minipage}[t]{0.9\linewidth}
    \includegraphics[width=0.99\linewidth]{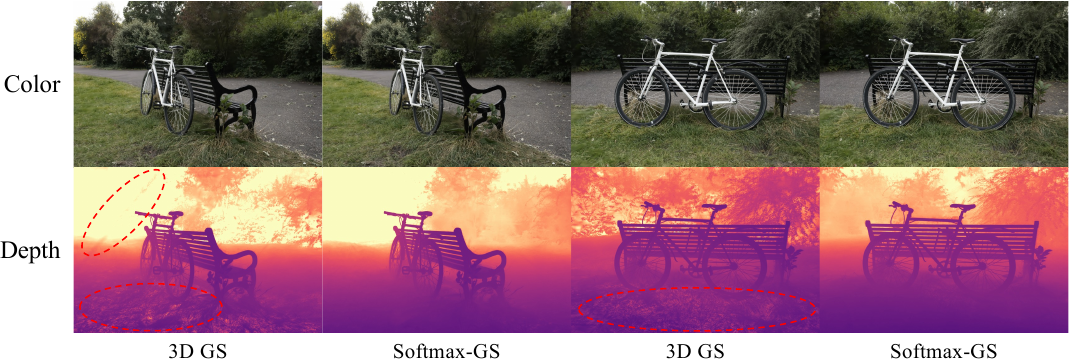}
    \vspace{-0.1in}
    \caption{\footnotesize Depth rendering comparison with 3D GS.}
    \label{fig:dep_bicycle}
    \end{minipage}
\end{figure}

We visualize the depth rendering of \methodname{} in Fig.~\ref{fig:dep_synth} for a synthetic pattern and in Fig.~\ref{fig:dep_bicycle} for a real-world scene. Note the smooth depth transitions at Gaussian boundaries.

\section{Non-coplanar intersection} 
\begin{figure}
    \centering
    \includegraphics[width=0.99\linewidth]{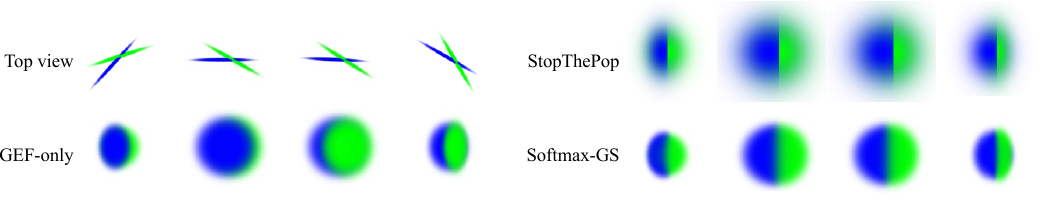}
    \vspace{-0.2in}
    \caption{\footnotesize Two Gaussians crossing at 30$^\circ$ angle.}
    \label{fig:cross}
\end{figure}

We visualize two Gaussians crossing at 30$^\circ$ in Fig.~\ref{fig:cross}. In contrast to the popping artifacts of 3D GS and the abrupt color changes of STP, \methodname{} produces smooth, flicker-free transitions at the crossing. 
Note \methodname{} effectively merges intersecting Gaussians into a single surface, consistent with the physical assumption that real-world object surfaces do not cross.

\setlength{\tabcolsep}{3pt}
\begin{table*}[htb]
\centering
\begin{minipage}[t]{0.9\linewidth}
\begin{center}
    \captionsetup{type=table}
    \hspace{-0.1in}\resizebox{\linewidth}{!}{
\begin{scriptsize}
\renewcommand{\arraystretch}{1.2}
\begin{tabular}{@{}$l^c^c^c^c^c^c^c^c^c^c^c@{}}
\toprule
\rowstyle{\bfseries}
 \multirow{2}{*}{Method} &  \multicolumn{7}{^c}{Mip-NeRF360} & \multicolumn{2}{^c}{Tanks\&Temples} & \multicolumn{2}{^c}{Deep Blending} \\
 \cmidrule(lr){2-8}\cmidrule(lr){9-10}\cmidrule(lr){11-12}
\rowstyle{\ttfamily} & bicycle & bonsai & counter & garden & kitchen & room & stump & train & truck & drjohnson & playroom \\
 \midrule
3D GS & 25.13 (4.8M) & 32.45 (1.0M) & 29.26 (1.0M) & 27.29 (4.5M) & 31.42 (1.5M) & 31.79 (1.2M) & 26.75 (4.1M) & 22.19 (1.1M) & 25.12 (2.7M) & 28.72 (3.1M) & 30.18 (2.0M) \\
\methodname{} & \cellcolor{first}25.24 (4.6M) & \cellcolor{first}32.78 (1.0M) & \cellcolor{first}29.38 (1.0M) &\cellcolor{first} 27.41 (4.4M) & \cellcolor{first}32.11 (1.4M) & \cellcolor{first}32.18 (1.2M) & \cellcolor{first}26.78 (3.7M) &\cellcolor{first} 22.62 (1.1M) & \cellcolor{first}25.55 (2.6M) & \cellcolor{first}29.01 (2.9M) & \cellcolor{first}30.53 (1.9M) \\
\bottomrule
\end{tabular}
\end{scriptsize}}
\end{center}
\vspace{-0.3cm}
\caption{\footnotesize Per-scene PSNR and number of Gaussians (in bracket) comparison between 3D GS and \methodname{}.
}
\label{tb:smgs_real_indiv}
\end{minipage}
\end{table*}

\section{Per-scene results}
Per-scene comparisons of PSNR and Gaussian counts are presented in Table~\ref{tb:smgs_real_indiv}, showing that \methodname{} achieves higher rendering quality with a similar number of Gaussians across all scenes.

\section{Full Algorithm}
We provide the complete forward-pass of the \methodname{} algorithm in Algorithm~\ref{alg:smgs}.
\begin{algorithm*}
\caption{The \methodname{} rendering algorithm. For simplicity, we ignore invalid Gaussians and background color, and assume there are $K$ Gaussians in total.}\label{alg:smgs}
\begin{algorithmic}[1]
\Require $T_{\text{past}}=1, c_{\text{past}}=0$ \Comment{Transmittance and pixel color so far}
\Require $d_{\text{past}}=0, p_{\text{past}}=0$ \Comment{Moving average of past depth and Gaussian exponent}
\Require $K, \textbf{x}_{\text{pixel}}$ \Comment{Total number of Gaussians along the ray, pixel coordinate}
\Require $\textbf{x}[K], \mathbf{\sigma}[K],o[K],c[K],d[K]$ \Comment{Gaussian splat center, conics, opacity, color, depth}
\Require $\alpha[K],\beta[K], \gamma[K]$ \Comment{\methodname{} parameters}
\State $k\gets 1$
\While{$k \leq K$}
\State $\textbf{x}'=\textbf{x}[k]-\textbf{x}_{\text{pixel}}$
\State $p_{\text{cur}}\gets -0.5\cdot \text{Mahalanobis\_distance}(\textbf{x}',\sigma[k])$
\State $a_{\text{cur}} \gets o[k]\cdot \exp(-(-p_{\text{cur}})^{\alpha[k]})$ \Comment{Control boundary sharpness using GEF}

\If{$T_{\text{past}}<1$}
    \State $T_{\text{orig}}\gets T_{\text{past}}\cdot (1-{a}_{\text{cur}})$ \Comment{Record the original transmittance}
    \State $a_{\text{past}}\gets 1-T_{\text{past}}$\Comment{Past absorbance}
    \State $w_{\text{cur}}\gets 1 / (1+\exp(\beta[k] \cdot (p_{\text{past}}-p_{\text{cur}}))$\Comment{Softmax competition between the Gaussians}
    \State $\hat{a}_{\text{cur}}\gets w_{\text{cur}}\cdot a_{\text{cur}}$\Comment{Softmax-weighted current absorbance}
    \State $w_{\text{past}}\gets 1 - w_{\text{cur}}$
    \State $\hat{a}_{\text{past}}\gets w_{\text{past}}\cdot a_{\text{past}}$\Comment{Softmax-weighted past absorbance}
    \State $\hat{a}_{\text{past}}\gets \frac{\hat{a}_{\text{past}}(1-T_{\text{orig}})}{\hat{a}_{\text{past}}+\hat{a}_{\text{cur}}}$\Comment{Order invariance and transmittance maintenance}
    \State $\hat{a}_{\text{cur}}=\frac{\hat{a}_{\text{cur}}(1-T_{\text{orig}})}{\hat{a}_{\text{cur}}+ \hat{a}_{\text{past}}T_{\text{orig}}}$
    \State $s\gets \exp(-\gamma[k]|d[k]-d_\text{past}|)$\Comment{Decay influence of softmax with distance}
    \State $\bar{a}_{\text{past}} \gets s\cdot \hat{a}_{\text{past}}+(1-s)\cdot a_{\text{past}}$ \Comment{Interpolate original and softmax-ed absorbance}
    \State $a_{\text{cur}} \gets s\cdot \hat{a}_{\text{cur}}+(1-s)\cdot a_{\text{cur}}$
    \State $m\gets \frac{a_{\text{cur}}+\bar{a}_{\text{past}}-\sqrt{(a_{\text{cur}}+\bar{a}_{\text{past}})^2-4(1-T_{\text{orig}})a_{\text{cur}}\bar{a}_{\text{past}}}}{2a_{\text{cur}}\bar{a}_{\text{past}}}$ \Comment{Transmittance maintenance again}
    \State $\bar{a}_{\text{past}}\gets m\cdot \bar{a}_{\text{past}}$
    \State $a_{\text{cur}}\gets m\cdot a_{\text{cur}}$
    \State $T_{\text{past}}\gets 1 - \bar{a}_{\text{past}}$\Comment{Update past transmittance}
    \State $c_{\text{past}}\gets c_{\text{past}} \cdot \bar{a}_{\text{past}} / a_{\text{past}} $\Comment{Update past color}
\EndIf
\State $c_{\text{past}}\gets c_{\text{past}}+c[k]\cdot a_{\text{cur}}\cdot T_{\text{past}}$
\State $d_{\text{past}} \gets \frac{d_{\text{past}}\cdot (1-T_{\text{past}})+d[k]\cdot a_{\text{cur}}\cdot T_{\text{past}}}{1-T_{\text{past}}+a_{\text{cur}}\cdot T_{\text{past}}}$\Comment{Average depth weighted by absorbance}
\State $p_{\text{past}} \gets \frac{p_{\text{past}}\cdot (1-T_{\text{past}})+p_{\text{cur}}\cdot a_{\text{cur}}\cdot T_{\text{past}}}{1-T_{\text{past}}+a_{\text{cur}}\cdot T_{\text{past}}}$\Comment{Average Gaussian exponet weighted by absorbance}
\State $T_{\text{past}}\gets T_{\text{past}}\cdot (1-a_{\text{cur}})$
\State $k\gets k+1$
\EndWhile
\end{algorithmic}
\end{algorithm*}

\section{Limitations}
\methodname{} has three main limitations.
First, the proposed splatting algorithm is applied only to the first 128 Gaussians along each ray in order to maintain linear complexity in the backward pass. As a result, coverage is incomplete: on Mip-NeRF360 indoor scenes, \methodname{} accounts for approximately 85\% of pixels across test images, while for outdoor scenes the coverage drops to around 70\%.
Second, the order-invariance mechanism in \methodname{} does not extend to cases with three or more overlapping Gaussians with distinct colors. This limitation arises from the bookkeeping complexity required to preserve permutation invariance under a strict linear-time constraint.
Third, the current formulation struggles with semi-transparent Gaussians. In particular, distant semi-transparent Gaussians can bias the running estimates of accumulated depth and opacity toward intermediate values, which in turn affects the softmax-based competition among subsequent Gaussians along the ray.
Addressing these limitations is an important direction for future work.



\end{document}